\documentclass{article} 
\usepackage{nips12submit_e,times}

\usepackage[pdftex]{graphicx}

\usepackage{xspace}
\usepackage{bm}
\usepackage[square]{natbib}
\usepackage{amsmath,amssymb}
\usepackage{amsthm}
\usepackage{eqparbox}
\usepackage{caption}
\usepackage{subcaption} 
\usepackage{wrapfig}

\usepackage{algorithm}
\usepackage[noend]{algorithmic}

\usepackage{hyperref}

\newtheorem{theorem}{Theorem}

\newtheorem{proposition}[theorem]{Proposition}
\newtheorem{corollary}[theorem]{Corollary}
\usepackage[scaled]{helvet}

\newcommand{\X}{\mathbf{X}}
\newcommand{\w}{\mathbf{w}}

\newcommand{\y}{\mathbf{y}}

\newcommand{\RR}{I\!\!R} 

\newcommand{\data}[1]{{\sffamily \scshape #1}\xspace}

\newcommand{\inner}[1]{\left\langle {#1} \right\rangle}

\title{Feature Clustering for Accelerating\\Parallel Coordinate Descent}

\author{
Chad Scherrer \\
Independent Consultant\\
Yakima, WA  \\
\texttt{chad.scherrer@gmail.com} \\
\And
Ambuj Tewari \\
Department of Statistics\\
University of Michigan\\
Ann Arbor, MI\\
\texttt{tewaria@umich.edu} \\
\AND
Mahantesh Halappanavar\\
Pacific Northwest National Laboratory\\
Richland, WA \\
\texttt{mahantesh.halappanavar@pnnl.gov} \\
\And
David J Haglin \\
Pacific Northwest National Laboratory \\
Richland, WA \\
\texttt{david.haglin@pnnl.gov}
}

\nipsfinalcopy 

\DeclareMathOperator*{\argmin}{argmin}

\begin{document}

\maketitle

\begin{abstract} 
Large-scale $\ell_1$-regularized loss minimization problems arise in high-dimensional applications such as compressed sensing and high-dimensional supervised learning, including classification and regression problems. High-performance algorithms and implementations are critical to efficiently solving these problems. Building upon previous work on coordinate descent algorithms for $\ell_1$-regularized problems, we introduce a novel family of algorithms called block-greedy coordinate descent that includes, as special cases, several existing algorithms such as SCD, Greedy CD, Shotgun, and Thread-Greedy. We give a unified convergence analysis for the family of block-greedy algorithms. The analysis suggests that block-greedy coordinate descent can better exploit parallelism if features are clustered so that the maximum inner product between features in different blocks is small. Our theoretical convergence analysis is supported with experimental results using data from diverse real-world applications. We hope that algorithmic approaches and convergence analysis we provide will not only advance the field, but will also encourage researchers to systematically explore the design space of algorithms for solving large-scale $\ell_1$-regularization problems. 
\end{abstract}

\section{Introduction}

Consider the $\ell_1$-regularized loss minimization problem
\begin{equation}
\label{eq:l1lossmin}
\min_{\w} \
 \frac{1}{n} \sum_{i=1}^n \ell(\y_i,(\X\w)_i)
+ \lambda \|\w\|_1\ ,
\end{equation}
where $\X\in\RR^{n\times p}$ is the design matrix, $\w\in\RR^p$ is a weight vector to be estimated, and the loss function $\ell$ is such that $\ell(y,\cdot)$ is a convex differentiable function for each $y$.
This formulation includes $\ell_1$-regularized least squares (Lasso) (when $\ell(y,t) = \tfrac{1}{2} (y-t)^2$) and $\ell_1$-regularized logistic regression 
(when $\ell(y,t) = \log(1+\exp(-yt))$).
In recent years, coordinate descent (CD) algorithms have been shown to be efficient for this class of problems \citep{Friedman2007,Wu2008,Shalev-Shwartz2011,Bradley2011}.

Motivated by the need to solve large scale $\ell_1$ regularized problems, researchers have begun to explore parallel algorithms. For instance, \citet{Bradley2011}
developed the Shotgun algorithm. More recently, \cite{Anon2012} have developed ``GenCD'', a generic framework for expressing parallel coordinate descent algorithms.
Special cases of GenCD include Greedy CD \citep{Li2009,Dhillon2011}, the Shotgun algorithm of \citep{Bradley2011}, and Thread-Greedy CD \citep{Anon2012}.


In fact, the connection between these three special cases of GenCD is much deeper, and more fundamental, than is obvious under the GenCD abstraction. As our first contribution, we describe a general randomized \emph{block-greedy} that includes all three as special cases. The block-greedy algorithm has two parameter: $B$, the total number of feature blocks and $P$, the size of the \emph{random subset} of the $B$ blocks that is chosen at every time step. For each of these $P$ blocks, we greedily choose, in parallel, a single
feature weight to be updated.

Second, we present a non-asymptotic convergence rate analysis for the randomized block-greedy coordinate descent algorithms for general values of $B \in \{1,\ldots,p\}$ (as the
number of blocks cannot exceed the number of features) and $P \in \{1,\ldots,B\}$. This result therefore applies
to stochastic CD, greedy CD, Shotgun, and thread-greedy. Indeed, we build on the analysis and insights in all of these previous works.
Our general convergence result, and in particular its instantiation to thread-greedy CD, is novel.

Third, based on the convergence rate analysis for block-greedy, we optimize a certain ``block spectral radius" associated with the design matrix.
This parameter is a direct generalization of a similar spectral parameter that appears in the analysis of Shotgun. We show that the block spectral radius
can be upper bounded by the maximum inner product (or correlation if features are mean zero) between features in distinct blocks. This motivates the use of
correlation-based feature clustering to accelerate the convergence of the thread-greedy algorithm.

Finally, we conduct an experimental study using a simple clustering heuristic. We observe dramatic acceleration due to clustering for smaller values of the regularization parameter, and show characteristics that must be paid particularly close attention for heavily regularized problems, and that can be improved upon in future work.


\section{Block-Greedy Coordinate Descent}

%
%

\cite{Anon2012} describe ``GenCD'', a generic framework for parallel coordinate descent algorithms, in which a parallel coordinate descent algorithm can be determined by specifying a \emph{select} step and an \emph{accept} step. At each iteration, features chosen by \emph{select} are evaluated, and a proposed increment is generated for each corresponding feature weight. Using this, the \emph{accept} step then determines which proposals are to be updated.

\begin{wrapfigure}{R}{1.5in}
\centering
\vspace{-2em}
\includegraphics[width=1.48in]{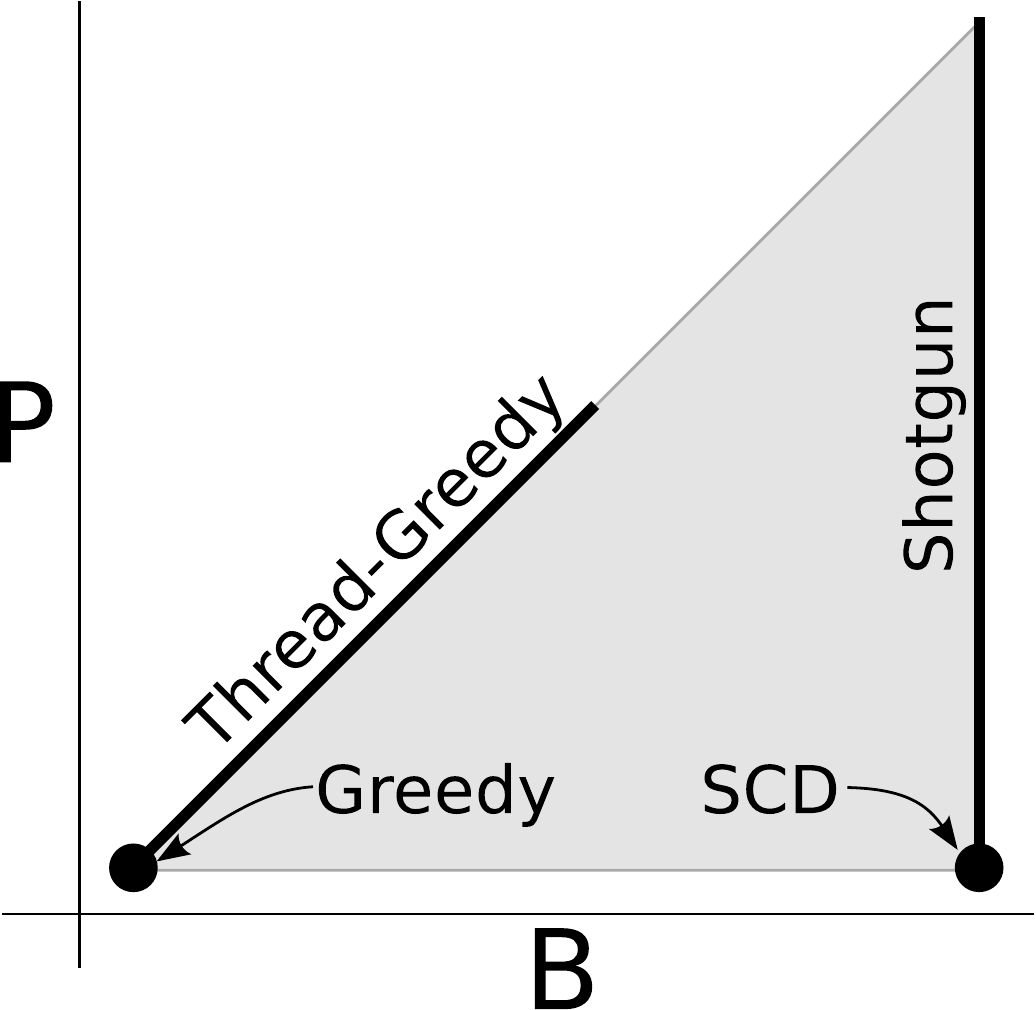}
\caption{The design space of block-greedy coordinate descent.}
\label{fig:algorithms-schematic}
\end{wrapfigure}

In these terms, we consider the \emph{block-greedy} algorithm that takes as part of the input a partition of the features into $B$ \emph{blocks}. Given this, each iteration \emph{select}s features corresponding to a set of $P$ randomly selected blocks, and \emph{accept}s a single feature from each block, based on an estimate of the resulting reduction in the objective function.


The pseudocode for the randomized block-greedy coordinate descent is given by Algorithm~\ref{alg:block-greedy}. The algorithm
can be applied to any function of the form $F+R$ where $F$ is smooth and convex, and $R$ is convex and separable across coordinates. 
Our objective function~\eqref{eq:l1lossmin} satisfies these conditions. The greedy step chooses
a feature within a block for which the \emph{guaranteed descent} in the objective function (if that feature alone were updated) is maximized.
This descent is quantified by $|\eta_j|$, which is defined precisely in the next section. To arrive at an heuristic understanding,
it is best to think of $|\eta_j|$ as being proportional to the absolute value $|\nabla_j F(\w)|$ of the $j$th entry in the gradient of the
smooth part $F$. In fact, if $R$ is zero (no regularization) then this heuristic is exact.

\begin{algorithm}
\caption{Block-Greedy Coordinate Descent}
\label{alg:block-greedy}
\begin{algorithmic}
\STATE {\bf Parameters:} $B$ (no. of blocks) and $P \le B$ (degree of parallelism)
\WHILE{not converged}
\STATE {\emph{Select} a random subset of size $P$ from the $B$ available blocks}
\STATE {Set $J$ to be the features in the selected blocks}
\STATE {\emph{Propose} increment $\eta_j, j\in J$} \COMMENT{parallel}
\STATE {\emph{Accept} $J' = \{j \::\: \eta_j \text{ has maximal absolute value in its block}\}$}
\STATE {\emph{Update} weight $w_j \gets w_j - \eta_j$ for all $j\in J'$} \COMMENT{parallel}
\ENDWHILE
\end{algorithmic}
\end{algorithm}

The two parameters, $B$ and $P$, of the block-greedy CD algorithm have
the ranges $B \in \{1,\ldots,p\}$ and $P \in \{1,\ldots,B\}$. Setting these to specific values gives many existing algorithms.
For instance when $B = p$, each feature is a block on its own. Then, setting $P = 1$ amounts to
randomly choosing a single coordinate and updating it which gives
us the stochastic CD algorithm of \citet{Shalev-Shwartz2011}.
Shotgun \citep{Bradley2011} is obtained when $B$ is still $p$ but $P \ge 1$.
Another extreme is the case when all the features constitute a single
block. That is, $B = 1$. Then block-greedy CD is a deterministic algorithm and becomes the greedy CD algorithm of \cite{Li2009,Dhillon2011}.
Finally, we can choose non-trivial values of $B$ that lie strictly between $1$ and $p$. When this is the case, and we
choose to update all blocks in parallel each time ($P = B$), we arrive at the thread-greedy algorithm of \citet{Anon2012}.
Figure~\ref{fig:algorithms-schematic} shows a schematic representation of the parameterization of these special cases.




\section{Convergence Analysis}\label{sect:convergence}

\def\rhoblock{\rho_{\mathrm{block}}}

Of course, there is no reason to expect block-greedy CD to converge for all values of $B$ and $P$. In this section, we give a sufficient condition for convergence and derive a convergence rate assuming this condition.

 \citeauthor{Bradley2011} express the convergence criteria for Shotgun algorithm in terms of the spectral radius (maximal eigenvalue) $\rho(\X^T \X)$. For block-greedy, the corresponding quantity is a bit more complicated. We define
\[
	\rhoblock = \max_{M \in \mathcal{M}} \rho(M) \ 
\]
where $\mathcal{M}$ is the set of all $B \times B$ submatrices that we can obtain from $\X^T \X$ by selecting exactly one index from each of the $B$ blocks.
The intuition is that if features from different blocks are almost orthogonal then the matrices $M$ in $\mathcal{M}$ will be close to identity and will therefore
have small $\rho(M)$. Highly correlated features \emph{within} a block do not increase $\rhoblock$.

\def\reals{\mathbb{R}}
\newcommand\E[1]{\mathbb{E}\left[{#1}\right]}
\newcommand\Es[2]{\mathbb{E}_{#1}\left[{#2}\right]}

As we said above, we will assume that we are minimizing a ``smooth plus separable" convex function $F+R$ where the convex differentiable function
$F:\reals^p \to \reals$ satisfies a second order upper bound
\[
	F(\w + \Delta) \le F(\w) + \nabla F(\w)^T \Delta + \frac{\beta}{2} \Delta^T \X^T \X \Delta
\]
In our case, this inequality will hold as soon as $\ell''(y,t) \le \beta$ for any $y,t$ (where differentiation is w.r.t. $t$). The function $R$ is separable across
coordinates: $R(\w) = \sum_{j=1}^p r(w_j)$. The function $\lambda\|\w\|_1$ is clearly separable.

The quantity $\eta_j$ appearing in Algorithm~\ref{alg:block-greedy} serves to quantify the guaranteed descent (based on second order upper bound) if feature $j$ alone is
updated and is obtained as a solution of the one-dimensional minimization problem
\[
	\eta_j = \argmin_{\eta}\ \nabla_j F(\w) \eta + \frac{\beta}{2} \eta^2 + r(w_j+\eta) - r(w_j) \ .
\]
Note that if there is no regularization, then $\eta_j$ is simply $-\nabla_j F(\w)/\beta = -g_j/\beta$ (if we denote $\nabla_j F(\w)$ by $g_j$ for brevity).
In the general case, by first order optimality conditions for the above one-dimensional convex optimization problem, we have
$
	g_j + \beta \eta_j + \nu_j  = 0
$
where $\nu_j$ is a subgradient of $r$ at $w_j + \eta_j$. That is, $\nu_j \in \partial r(w_j + \eta_j)$. This implies that
$
	r(w_j + \eta_j) - r(w') \le \nu_j(w_j + \eta_j - w')
$
for any $w'$.

\begin{theorem}
\label{thm:main}
Let $P$ be chosen so that
\[
	\epsilon = \frac{(P-1)(\rhoblock-1)}{(B-1)}
\]
is less than $1$. Suppose the randomized block-greedy coordinate algorithm is run on a smooth plus separable convex function $f=F+R$ to produce the iterates
$\{\w_k\}_{k \ge 1}$. Then the expected accuracy after $k$ steps is bounded as
\[
	\E{ f(\w_k) - f(\w^\star) } \le C \frac{B\,R_{1}^2}{(1-\epsilon)P} \cdot \frac{1}{k} \ .
\]
Here the constant $C$ only depends on (Lipschitz and smoothness constants of) the function $F$, $R_{1}$ is an upper bound on
the norms $\{\|\w_k - \w^\star\|_{1}\}_{k \ge 1}$, and $\w^\star$ is any minimizer of $f$. 
\end{theorem}
\begin{proof}
We first calculate the expected change in objective function following the Shotgun analysis. We will use $w_b$ to denote $w_{j_b}$ (similarly for $\nu_b$, $g_b$ etc.)
\begin{align*}
\E{ f(\w') - f(\w) } &= P \Es{b}{ \eta_b g_b + \frac{\beta}{2} (\eta_b)^2 + r(w_b + \eta_b) - r(w_b)} \\
&\quad + \frac{\beta}{2} P(P-1) \Es{b \neq b'}{ \eta_b \cdot \eta_{b'} \cdot A_{j_b}^T A_{j_{b'}} } 
\end{align*}

Define the $B \times B$ matrix $M$ (that depends on the current iterate $\w$) with entries $M_{b,b'} = A_{j_b}^T A_{j_b}$. Then, using
$r(w_b + \eta_b) - r(w_b) \le \nu_b \eta_b$, we continue
\begin{align*}
&\le \frac{P}{B} \left[ \eta^T g + \frac{\beta}{2} \eta^T \eta + \nu^T \eta \right]
+\frac{\beta P(P-1)}{2 B (B-1)} \left[ \eta^\top M \eta - \eta^T \eta \right] 
\end{align*}
Above (with some abuse of notation), $\eta$, $\nu$ and $g$ are $B$ length vectors with components $\eta_b$, $\nu_b$ and $g_b$ respectively.
By definition of $\rhoblock$, we have $\eta^\top M \eta \le \rhoblock \eta^T \eta$.
So, we continue
\begin{align*}
&\le \frac{P}{B} \left[ \eta^T g + \frac{\beta}{2} \eta^T \eta - g^T \eta - \beta \eta^T \eta \right]
+\frac{\beta P(P-1)}{2 B (B-1)} (\rhoblock - 1) \eta^T \eta 
\end{align*}
where we used $\nu = -g - \beta \eta$.
Simplifying we get
\[
\E{f(\w') - f(\w)} \le \frac{P\beta}{2B} \left[ -1 + \epsilon \right] \|\eta\|_2^2
\]
where
\[
	\epsilon = \frac{(P-1)(\rhoblock-1)}{(B-1)}
\]
should be less than $1$.

Now note that
$\|\eta\|_2^2 = \sum_{b} |\eta_{j_b}|^2 = \| \eta \|_{\infty,2}^2$
where the ``infinity-2'' norm $\|\cdot\|_{\infty,2}$ of a $p$-vector is, by definition, as follows: take the $\ell_\infty$ norm within a block
and take the $\ell_2$ of the resulting values. Note that in the second step above, we moved from a $B$-length $\eta$ to a $p$ length $\eta$.

This gives us
\[
	\E{f(\w') -f(\w)} \le -\frac{(1-\epsilon)P\beta}{2 B} \| \eta \|_{\infty,2}^2 \ .
\]

For the rest of the proof, assume $\lambda = 0$. In that case $\eta = -g/\beta$. Thus,
convexity and the fact that the dual norm of the ``infinity-2" norm is the ``1-2" norm, give
\[
f(\w) - f(\w^\star) \le \nabla f(\w)^T(\w-\w^\star) \le \|\nabla f(\w)\|_{\infty,2} \cdot \|\w-\w^\star\|_{1,2}
\]
Putting the last two inequalities together gives (for any upper bound $R_{1}$ on $\|\w-\w^\star\|_1 \ge \|\w-\w^\star\|_{1,2}$)
\[
	\E{f(\w') - f(\w)} \le -\frac{(1-\epsilon)P}{2\beta B R_{1}^2 } (f(\w)-f(\w^\star))^2 \ .
\]
Defining the accuracy $\alpha_k = f(\w_k) - f(\w^\star)$, we translate the above into the recurrence
\[
	\E{ \alpha_{k+1} - \alpha_{k} } \le -\frac{(1-\epsilon)P}{2\beta B R_{1}^2 } \E{\alpha_k^2}
\]
and by Jensen's we have $(\E{\alpha_k})^2 \le \E{\alpha_k^2}$ and therefore
\[
	\E{\alpha_{k+1}} - \E{\alpha_{k}} \le -\frac{(1-\epsilon)P}{2\beta B R_{1}^2 } (\E{\alpha_k})^2
\]
which solves to (up to a universal constant factor)
\[
	\E{\alpha_k} \le \frac{2\beta B R_{1}^2}{(1-\epsilon)P} \cdot \frac{1}{k}
\]
Even when $\lambda > 0$, we can still relate $\|\eta\|_{\infty,2}$ to $f(\w) - f(\w^\star)$ but the argument is a little more involved.
We refer the reader to the supplementary for more details.
\end{proof}

In particular, consider the case where all blocks are updated in parallel as in the thread-greedy coordinate descent algorithm
of~\citet{Anon2012}. Then $P=B$ and there is no randomness in the algorithm, yielding the following corollary.

\begin{corollary}
Suppose the block-greedy coordinate algorithm with $B = P$ (thready-greedy) is run on a smooth plus separable convex function $f=F+R$ to produce the iterates
$\{\w_k\}_{k \ge 1}$. If $\rhoblock < 2$, then 
\[
	f(\w_k) - f(\w^\star) = O\left( \frac{1}{(2-\rhoblock) k} \right) \ .
\] 
\end{corollary}

\section{Feature Clustering}

The convergence analysis of section~\ref{sect:convergence} shows that we need to minimize the block spectral radius. Directly finding a clustering that minimizes $\rhoblock$ is a computationally daunting task. 
Even with equal-sized blocks, the number of possible partitions is $p!/\left(\frac{p}{B}\right)^B$.
In the absence of an efficient search strategy for this enormous space, we find it convenient to work instead in terms of the inner product of features from distinct blocks.
The following proposition makes the connection between these approaches precise.

\begin{proposition}
\label{prop:correlation}
Let $S\in\reals^{B\times B}$ be positive semidefinite, with $S_{ii}=1$, and $\left|S_{ij}\right|<\varepsilon$
for $i\neq j$. Then the spectral radius of $S$ has the upper bound
\[
\rho(S)\leq 1+\left(B-1\right)\varepsilon\ .
\]
\end{proposition}

\begin{proof}
Let $x$ be the eigenvector corresponding to the largest eigenvalue of $S$, scaled so that $\Vert x \Vert_1 = 1$. Then
\[
\rho\left(S\right)
  =  \left\Vert Sx\right\Vert _{1}
  = \sum_{i}\left|x_{i}+S_{ij}\sum_{j\neq i}x_{j}\right|
  \leq  \sum_{i}\left(\left|x_{i}\right|+\varepsilon\sum_{j\neq i}\left|x_{j}\right|\right)
  = 1+\left(B-1\right)\varepsilon
 \]
\end{proof}

Proposition~\ref{prop:correlation} tells us that we can partition the features into clusters using a heuristic approach that strives to minimize the maximum absolute
inner product between the features (columns of the design matrix) $\X_i$ and $\X_j$ where $i$ and $j$ are features in different blocks.

\subsection{Clustering Heuristic} 

Given $p$ features and $B$ blocks, we wish to distribute the features evenly among the blocks, attempting to minimize the absolute inner product
between features across blocks.
Moreover, we require an approach that is efficient, since any time spent clustering could instead have been used for iterations of the main algorithm.
We describe a simple heuristic that builds uniform-sized clusters of features.

To construct a given block, we select a feature as a ``seed'', and assign the nearest features to the seed (in terms of absolute inner product) to be in the same block. Because inner products with very sparse features result in a large number of zeros, we choose at each step the most dense unassigned feature as the seed. Algorithm~\ref{alg:clustering} provides a detailed description. This heuristic requires computation of $O(Bp)$ inner products. In practice it is very fast---less than three seconds for even the large \data{KDDA} dataset.

\begin{algorithm}
\caption{A heuristic for clustering $p$ features into $B$ blocks, based on correlation}
\label{alg:clustering}
\begin{algorithmic}
\STATE {$U \gets \{1,\cdots,p\}$}
\FOR{$b = 1$ to $B-1$}
\STATE {$s\gets \arg \max_{j\in U} \mathrm{NNZ}(\X_j)$}
\FOR[parallel]{$j\in U$}
\STATE $c_j\gets \left|\inner{\X_s, \X_j}\right|$
\ENDFOR
\STATE $J_b \gets \{j $ yielding the $ \lceil p/B\rceil $ largest values of  $ c_j \}$
\STATE {$U \gets U \backslash J_b$}
\ENDFOR
\STATE {$J_B \gets U$}
\RETURN $\{ J_b | b=1,\cdots,B\}$
\end{algorithmic}
\end{algorithm}

\section{Experimental Setup}

\textbf{Platform}
All our experiments are conducted on a $48$-core system
comprising of $4$ sockets and $8$ banks of memory.
Each socket is an AMD Opteron processor codenamed Magny-Cours,
which is a multichip processor with two $6$-core chips on a single die. 
Each $6$-core processor is equipped with a three-level memory
hierarchy as follows:
$(i)$ $64$ KB of \texttt{L1} cache for data and $512$ KB of \texttt{L2}
cache that are private to each core, and 
$(ii)$ $12$ MB of \texttt{L3} cache that is shared among the $6$
cores. Each $6$-core processor is linked to a $32$ GB memory bank with
independent memory controllers leading to a total system memory of
$256$ GB ($32 \times 8$) that can be globally addressed from each
core. 
The four sockets are interconnected using HyperTransport-3
technology\footnote{Further details on AMD Opteron can be found
at \url{http://www.amd.com/us/products/embedded/processors/opteron/Pages/opteron-6100-series.aspx}.}.

\textbf{Datasets}
A variety of datasets were chosen\footnote{from  \url{http://www.csie.ntu.edu.tw/~cjlin/libsvmtools/datasets/}} for experimentation; these are summarized in Table~\ref{tab:input-summary}. We consider four datasets:
$(i)$
\data{News20} contains about $20,000$ UseNet postings from $20$
newsgroups. The data was gathered by Ken Lang at Carnegie Mellon
University circa 1995.
$(ii)$
\data{Reuters} is the RCV1-v2/LYRL2004 Reuters text data described by
\citet{Lewis2004}. 
In this term-document matrix, each example is a training document, and
each feature is a term. Nonzero values of the matrix correspond to
term frequencies that are transformed using a standard tf-idf
normalization. 
$(iii)$
\data{RealSim} consists of about $73,000$ UseNet articles from four
discussion groups: simulated auto racing, simulated aviation, real
auto racing, and real aviation.
The data was gathered by Andrew McCallum while at Just Research circa
1997. 
We consider classifying real vs simulated data, irrespective of auto/aviation.
$(iv)$
\data{KDDA} represents data from the KDD Cup 2010 challenge on
educational data mining.
The data represents a processed version of the training set of the
first problem, algebra\_2008\_2009, provided by the winner from the
National Taiwan University. These four inputs cover a broad spectrum
of sizes and structural properties.
\begin{table}[t]\small
\centering
\begin{tabular}{|l|r|r|r|l|}
\hline 
{\bf Name} & {\bf \# Features} & {\bf \# Samples} & {\bf \# Nonzeros} &{\bf Source}\\ \hline
\hline
\data{News20}  & $1,355,191$ & $19,996$ & $9,097,916$ & \cite{Keerthi2005}\\ \hline

\data{Reuters} & $47,237$ & $23,865$ & $1,757,800$ & \cite{Lewis2004}\\ \hline

\data{RealSim} & $20,958$ & $72,309$ & $3,709,083$ & \cite{RealSim}\\ \hline

\data{KDDA} & $20,216,830$ & $8,407,752$ & $305,613,510$ & \cite{KDDA}\\ \hline

\end{tabular}
\caption{Summary of input characteristics.}
\label{tab:input-summary}
\end{table}











\textbf{Implementation}
For the current work, our empirical results focus on thread-greedy coordinate descent with 32 blocks. 
At each iteration, a given thread must step through the nonzeros of each of its features to compute the proposed increment (the $\eta_j$ of Section~\ref{sect:convergence})
and the estimated benefit of choosing that feature. Once this is complete, the thread (without waiting) enters the line search phase, where it remains until all threads are being updated by less than the specified tolerance. Finally, all updates are performed concurrently. We use OpenMP's atomic directive to maintain consistency.

\textbf{Testing framework}

 We compare the effect of clustering to randomization (i.e. features are randomly assigned to blocks), for a variety of values for the regularization parameter $\lambda$. To test the effect of clustering for very sparse weights, we first let $\lambda_0$ be the largest power of ten that leads to any nonzero weight estimates. This is followed by the next three consecutive powers of ten. For each run, we measure the regularized expected loss and the number of nonzeros at one-second intervals. Times required for clustering and randomization are negligible, and we do not report them here.

\section{Results}

\begin{figure*}[t]
\begin{subfigure}[b]{0.24\textwidth}
  \centering
  \includegraphics[width=\textwidth]{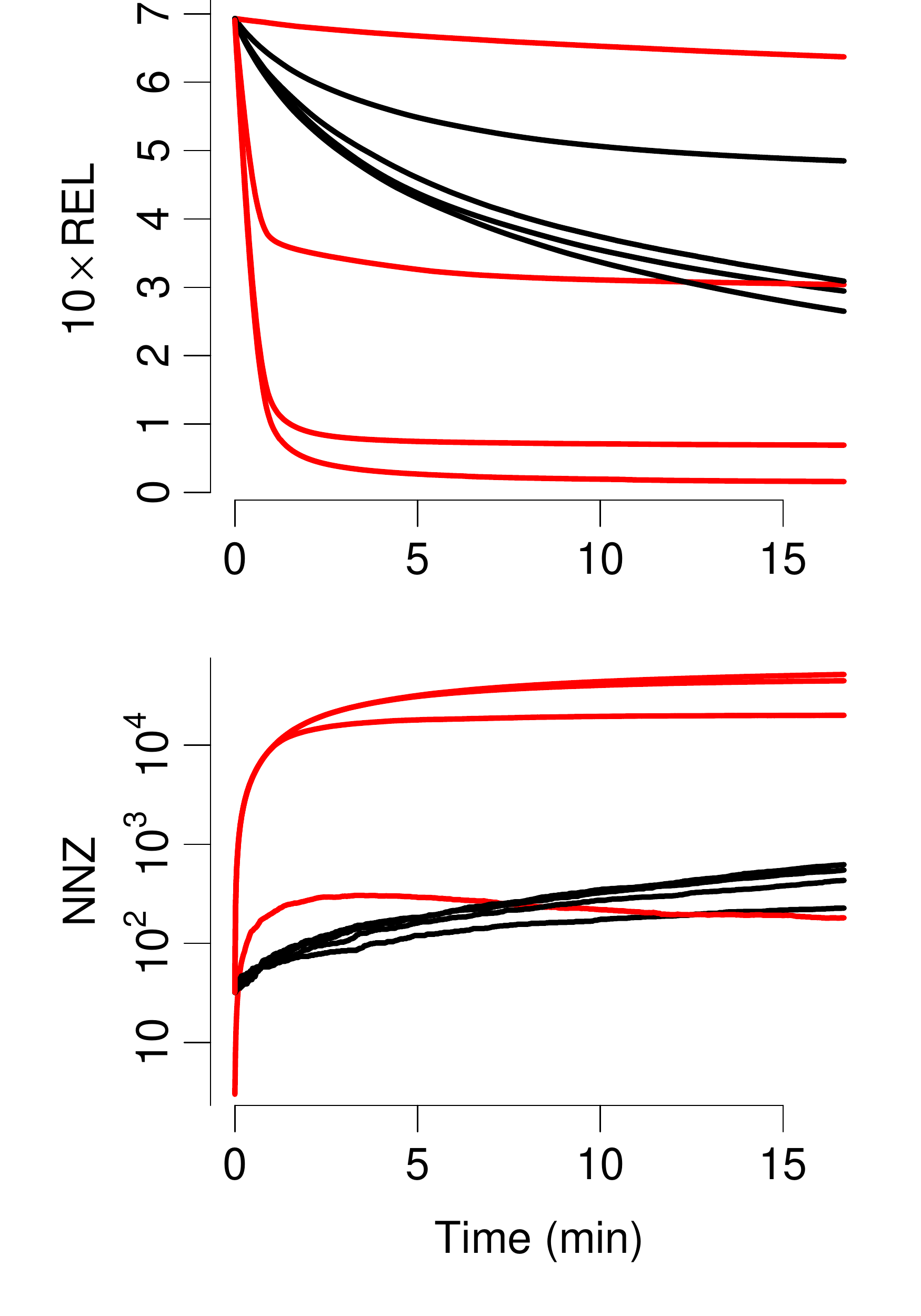}
  \caption{\data{News20}, \\$\lambda_0 = 10^{-4}$}
\end{subfigure}
\begin{subfigure}[b]{0.24\textwidth}
  \centering
  \includegraphics[width=\textwidth]{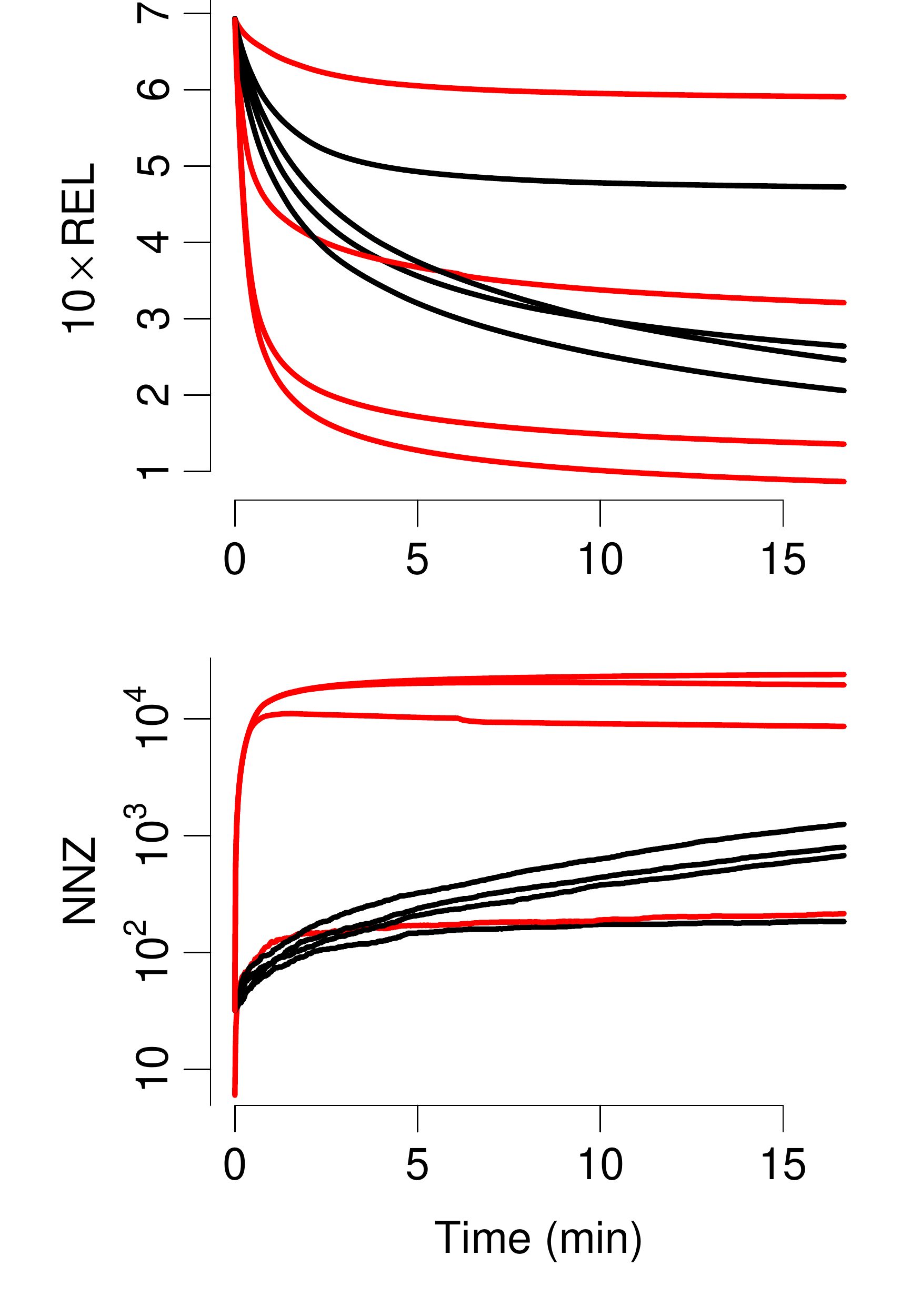}
  \caption{\data{Reuters}, \\$\lambda_0 = 10^{-4}$}
\end{subfigure}
\begin{subfigure}[b]{0.24\textwidth}
  \centering
  \includegraphics[width=\textwidth]{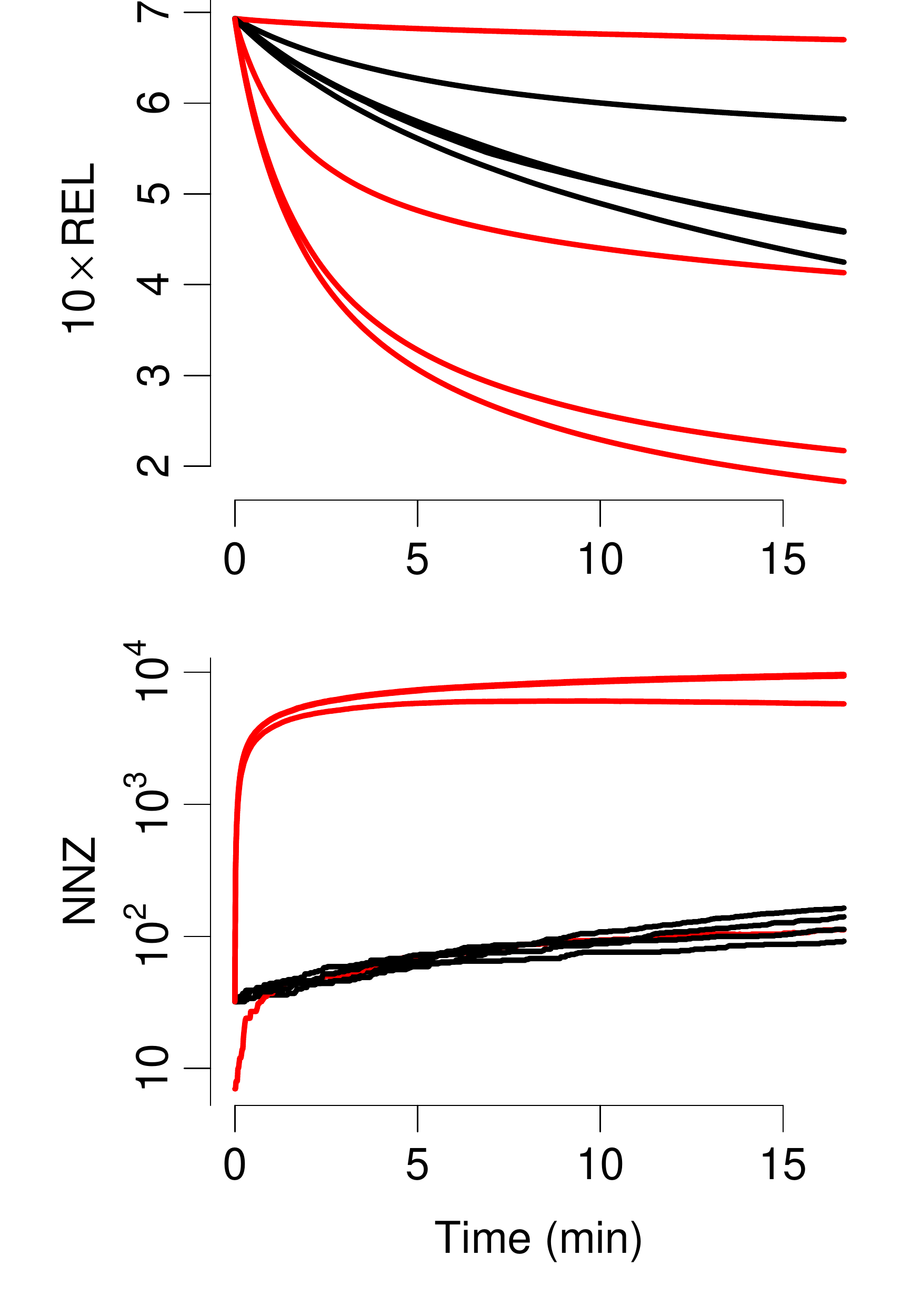}
  \caption{\data{RealSim}, \\$\lambda_0 = 10^{-4}$}
\end{subfigure}
\begin{subfigure}[b]{0.24\textwidth}
  \centering
  \includegraphics[width=\textwidth]{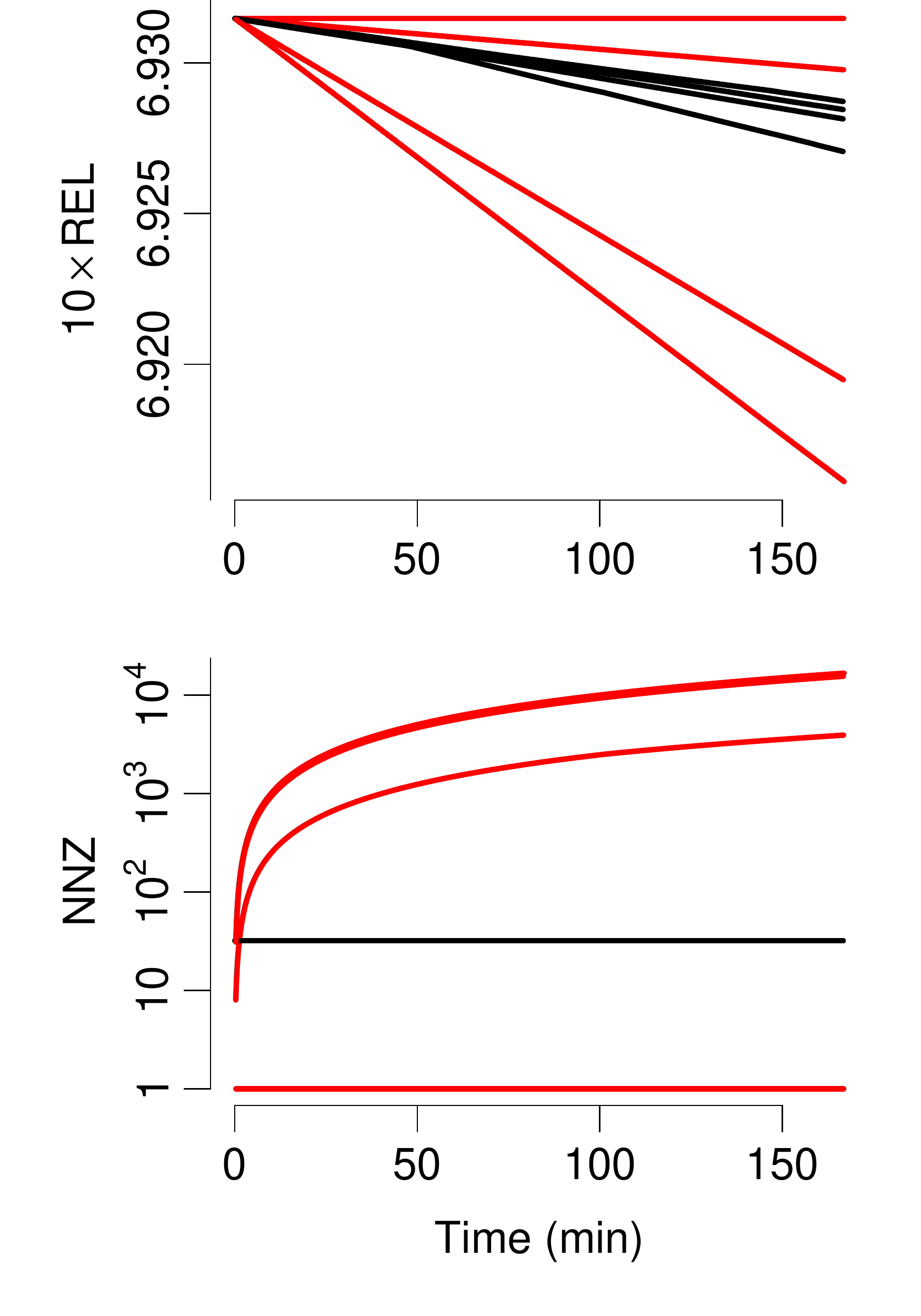}
  \caption{\data{KDDA}, \\$\lambda_0 = 10^{-6}$}
\end{subfigure}
\caption{Convergence results. For each dataset, we show the regularized expected loss (top) and number of nonzeros (bottom), using powers of ten as regularization parameters. Results for randomized features are shown in black, and those for clustered features are shown in red. Note that the allowed running time for \data{KDDA} was ten times that of other datasets.}
\label{fig:results}
\end{figure*}

Figure~\ref{fig:results} shows the regularized expected loss (top) and number of nonzeros (bottom), for several values of the regularization parameter $\lambda$. 
Black and red curves indicate randomly-permuted features and clustered features, respectively.
The starting value of $\lambda$ was $10^{-4}$ for all data except \data{KDDA}, which required $\lambda=10^{-6}$ in order to yield any nonzero weights.

In the upper plots, within a color, the order of the 4 curves, top to bottom, corresponds to successively decreasing values of $\lambda$.
Note that a larger value of $\lambda$  results in a sparser solution with greater regularized expected loss and a smaller number of nonzeros. 
Thus, for each subfigure of Figure~\ref{fig:results}, the order of the curves in the lower plot is \emph{reversed} from that of the upper plot. 

Overall, results across datasets are very consistent. For large values of $\lambda$, the simple clustering heuristic results in slower convergence, while for smaller values of $\lambda$ we see considerable benefit. Due to space limitations, we choose a single dataset for which to explore results in greater detail.

Of the datasets we tested, \data{Reuters} might reasonably lead to the greatest concern. Like the other datasets, clustered features lead to slow convergence for large $\lambda$ and fast convergence for small $\lambda$. However, \data{Reuters} is particularly interesting because for $\lambda=10^{-5}$, clustered features seem to provide an initial benefit that does not last; after about 250 seconds it is overtaken by the run with randomized features. 

\begin{table}[t]\small
\centering
\begin{tabular}{|c|cc|cc|cc|}
\hline 
 & \multicolumn{2}{c|}{$\lambda=10^{-4}$} &\multicolumn{2}{c|}{$\lambda=10^{-5}$} & \multicolumn{2}{c|}{$\lambda=10^{-6}$}\tabularnewline
\hline 
 & Randomized & Clustered & Randomized & Clustered & Randomized & Clustered\tabularnewline
\hline 
\hline 
Active blocks & {\bf 32} & 6 & 32 & 32 & 32 & 32\tabularnewline
Iterations per second & {\bf 153} & 12.9 & {\bf 152} & 12.3 & {\bf 136} & 12.3\tabularnewline
\hline 
NNZ @ 1K sec & 184 & 215 & 797 & 8592 & 1248 & 19473  \tabularnewline
Objective @ 1K sec & {\bf 0.472} & 0.591 & {\bf 0.264} & 0.321 & 0.206 & {\bf 0.136} \tabularnewline
\hline 
NNZ @ 10K iter & 74 & 203 & 82 & 8812 & 110 & 19919 \tabularnewline
Objective @ 10K iter & {\bf 0.570} & 0.593 & 0.515 & {\bf 0.328} & 0.472 & {\bf 0.141} \tabularnewline
\hline
\end{tabular}
\caption{The effect of feature clustering, for \data{Reuters}.}
\label{tab:reuters-summary}
\end{table}

Table~\ref{tab:reuters-summary} gives a more detailed summary of the results for \data{Reuters}, for the three largest values of $\lambda$. The first row of this table gives the number of \emph{active blocks}, by which we mean the number of blocks containing any nonzeros. For an inactive block, the corresponding thread repeatedly confirms that all weights remain zero without contributing to convergence. 

In the most regularized case $\lambda=10^{-4}$, clustered data results in only six active blocks, while for other cases every block is active. Thus in this case features corresponding to nonzero weights are colocated within these few blocks, severely limiting the advantage of parallel updates.

In the second row, we see that for randomized features, the algorithm is able to get through over ten times as many iterations per second. To see why, note that the amount of work for a given thread is a linear function of the number of nonzeros over all of the features in its block. Thus, the block with the greatest number of nonzeros serves as a bottleneck. 

The middle two rows of Figure~\ref{tab:reuters-summary} summarize the state of each run after 1000 seconds. Note that for this test, randomized features result in faster convergence for the two largest values of $\lambda$.

For comparison, the final two rows of Figure~\ref{tab:reuters-summary} provide a similar summary based instead on the number of iterations. In these terms, clustering is advantageous for all but the largest value of $\lambda$.

\begin{figure*}[t]
\begin{subfigure}[b]{0.33\textwidth}
  \centering
  \includegraphics[width=\textwidth]{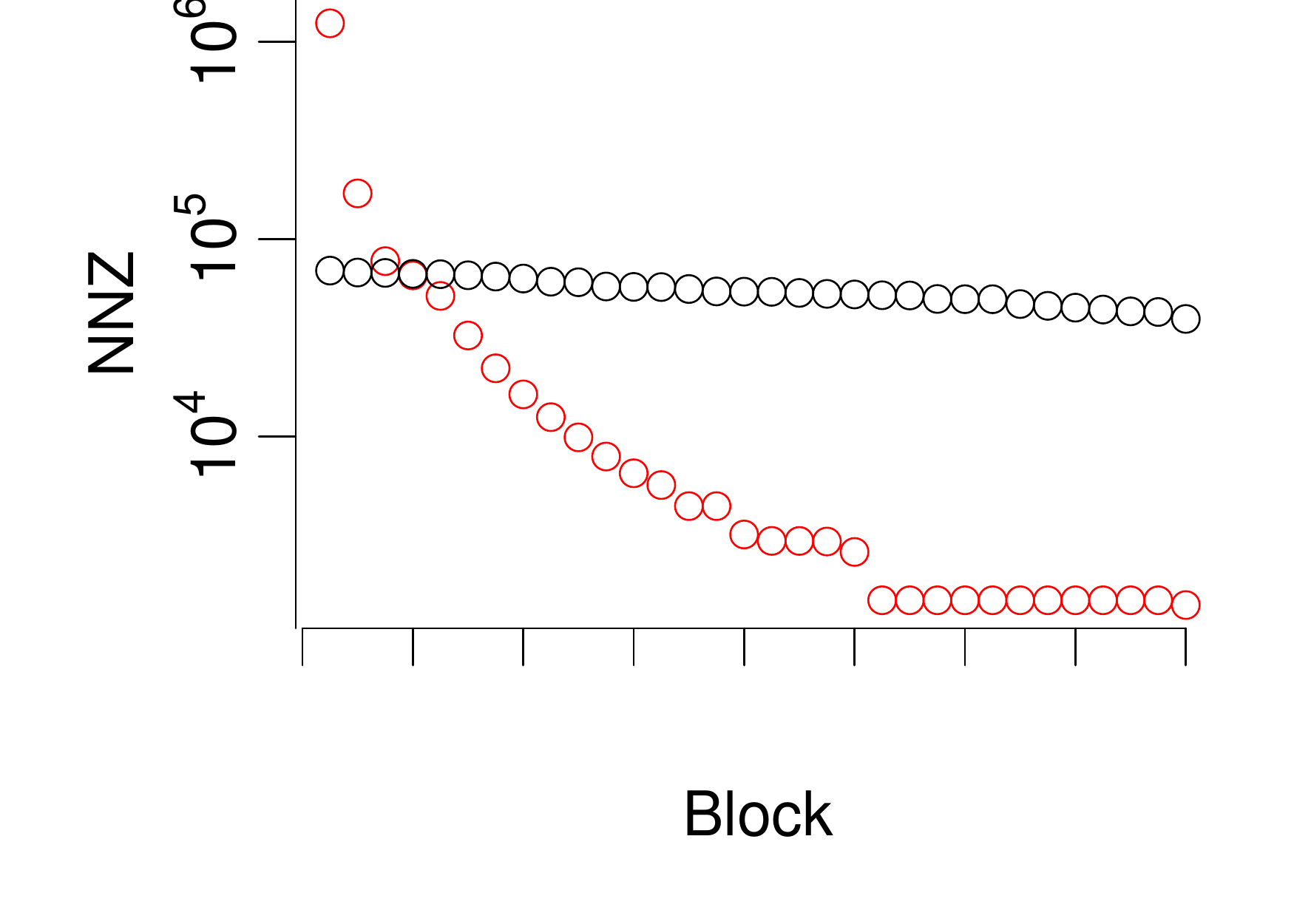}
  \caption{Block density}
  \label{fig:reut-blocknnz}
\end{subfigure}
\begin{subfigure}[b]{0.33\textwidth}
  \centering
  \includegraphics[width=\textwidth]{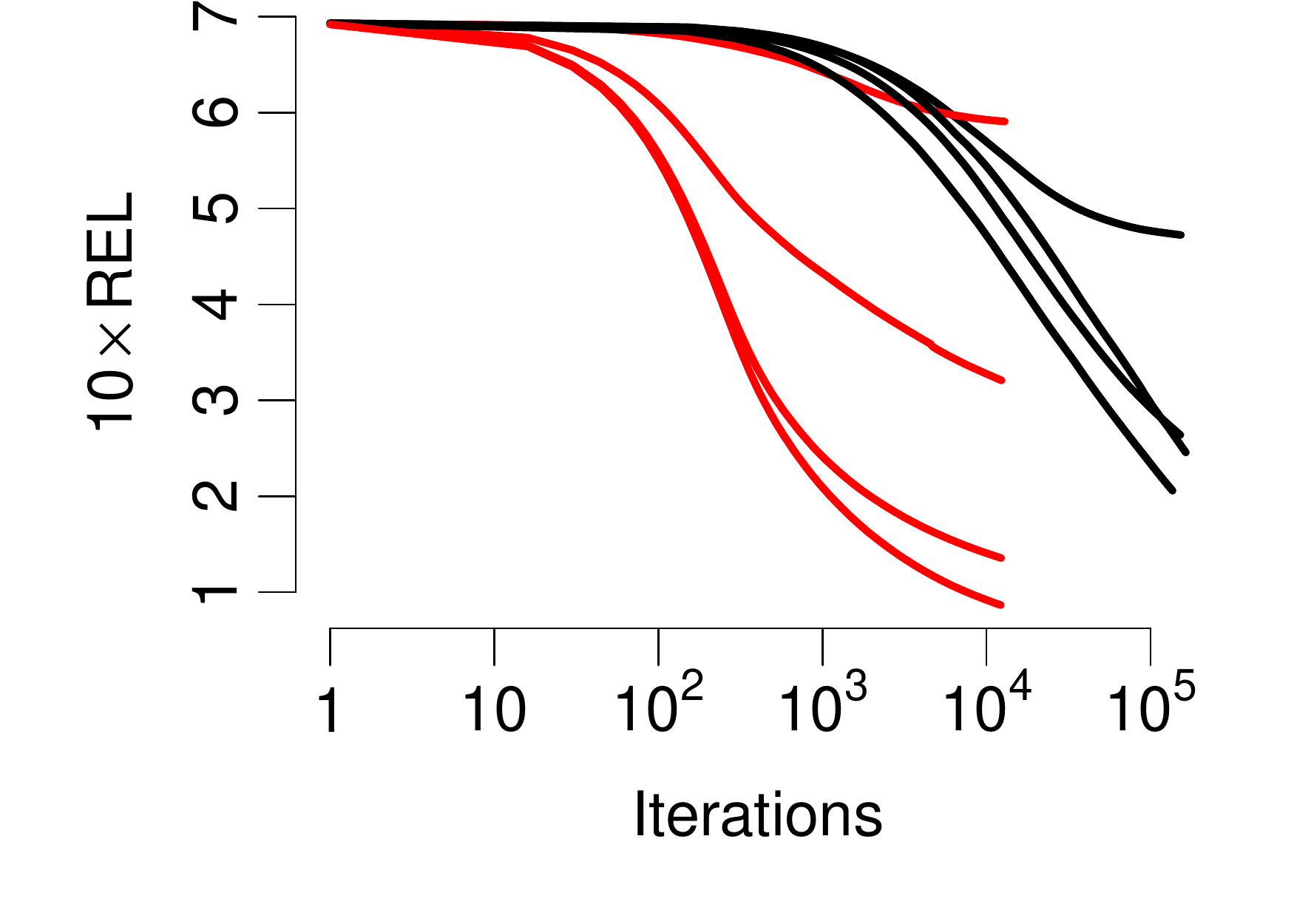}
  \caption{Regularized expected loss}
  \label{fig:reut-iter}
\end{subfigure}
\begin{subfigure}[b]{0.33\textwidth}
  \centering
  \includegraphics[width=\textwidth]{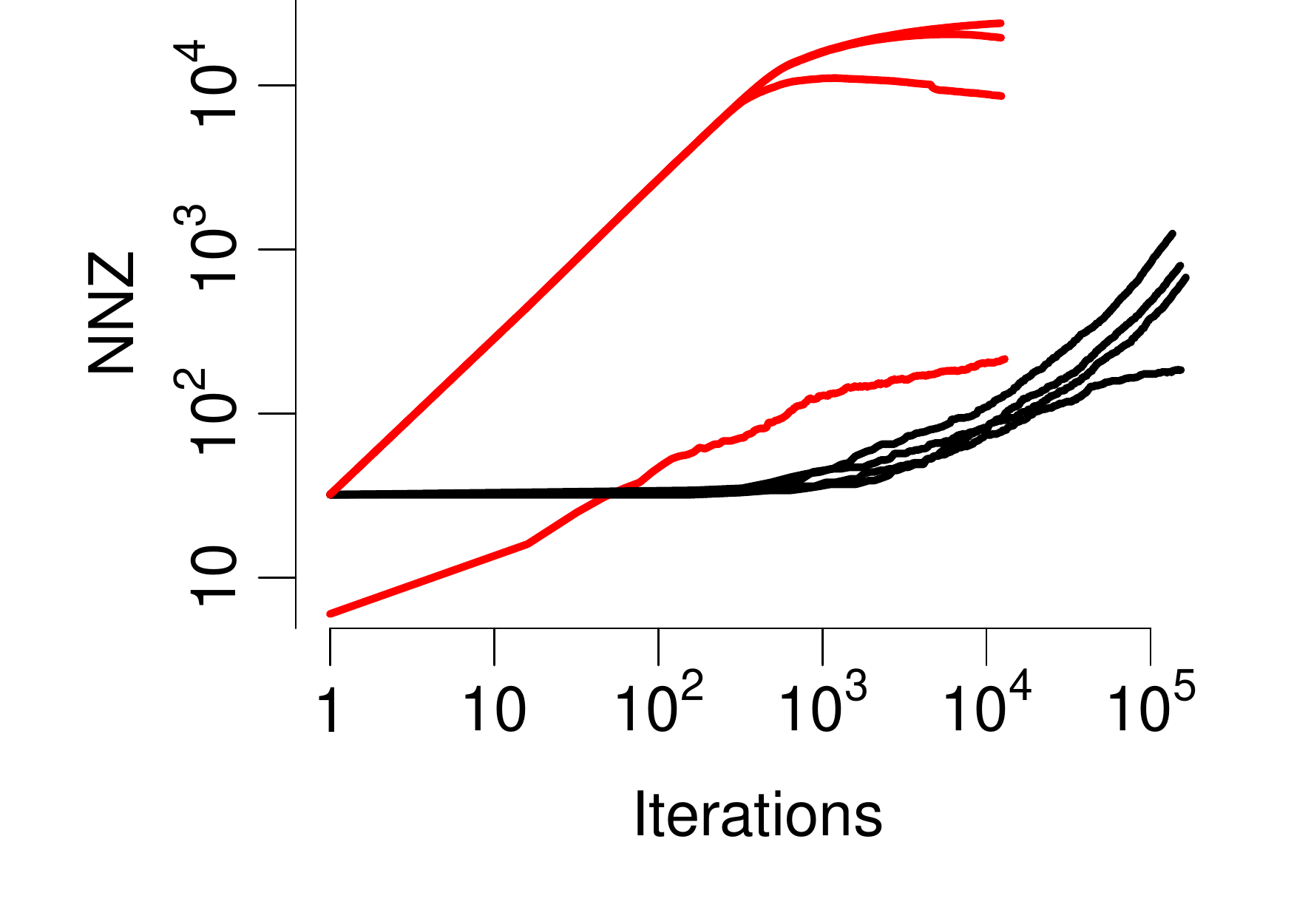}
  \caption{Number of nonzeros}
  \label{fig:reut-iter-nnz}
\end{subfigure}
\caption{A closer look at performance characteristics for \data{Reuters}.}
\label{fig:clustering-results}
\end{figure*}


Figure~\ref{fig:clustering-results} shows the source of this problem. First, Figure~\ref{fig:reut-blocknnz} shows the number of nonzeros in all features for each of the 32 blocks. Clearly the simple heuristic results in poor load-balancing. For comparison, Figures \ref{fig:reut-iter} and \ref{fig:reut-iter-nnz} show convergence rates as a function of the number of iterations.

\section{Conclusion}

We have presented convergence results for a family of randomized coordinate descent algorithms that we call \emph{block-greedy coordinate descent}. This family includes Greedy CD, Thread-Greedy CD, Shotgun, and Stochastic CD. We have shown that convergence depends on $\rhoblock$, the maximal spectral radius over submatrices of $\X^T \X$ resulting from the choice of one feature for each block.

Even though a simple clustering heuristic helps for smaller values of the regularization parameter, our results also show the importance of considering issues of load-balancing and the distribution of weights for heavily-regularized problems.

A clear next goal in this work is the development of a clustering heuristic that is relatively well load-balanced and distributes weights for heavily-regularized problems evenly across blocks, while maintaining good computational efficiency.

\subsubsection*{Acknowledgments}
The authors are grateful for the helpful suggestions of Ken Jarman, Joseph Manzano, and our anonymous reviewers.

Funding for this work was provided by the Center for Adaptive Super Computing Software - MultiThreaded Architectures (CASS-MT) at the U.S. Department of Energy's Pacific Northwest National Laboratory. PNNL is operated by Battelle Memorial Institute under Contract DE-ACO6-76RL01830. 


\bibliographystyle{unsrtnat}
\bibliography{main}


\newpage
\begin{center}
\Large
 \textbf{Supplementary Material}
\vspace{2em}
\end{center}
\appendix

\section{Complete convergence analysis in the regularized case}

\textbf{Basic setup}: We are minimizing a function $f$ of the form $F + R$ where $F$ is a convex differentiable function $F:\reals^p \to \reals$ that satisfies a second order
upper bound
\[
	F(w + \Delta) \le F(w) + \nabla F(w)^T w + \frac{\beta}{2} \Delta^T A^T A \Delta
\]
and $R:\reals^p \to \reals$ is convex (and possibly non-differentiable) and separable across coordinates:
\[
	R(w) = \sum_{j=1}^p r(w_j)
\] 
In our case $\X$ is the $n \times p$ design matrix. If columns of $\X$ are zero mean and unit variance normalized then entries in $\X^T \X$
measure the correlation between features. Also, $r(x) = \lambda |x|$.

Divide the $p$ features into $B$ blocks of $p/B$ features each. The algorithm we analyze is block-greedy, a direct generalization of
Shotgun ($B = p$ in the Shotgun case). In the regularized case, the block-greedy algorithm is

\noindent
\textbf{For} $P$ randomly chosen blocks in parallel \textbf{do}
\begin{itemize}
\item
Within a block $b$, find $j = j_b \in b$ such that $|\eta_j|$ is maximum and update
\[
w'_j \gets w_j - \eta_j
\]
\end{itemize}
\noindent
\textbf{Endfor}

Here $|\eta_j|$ serves to quantify the guaranteed descent (based on second order upper bound) if feature $j$ is updates and solves the one-dimensional problem
\[
	\eta_j = \argmin_{\eta}\ \nabla_j F(w) \eta + \frac{\beta}{2} \eta^2 + r(w_j+\eta) - r(w_j) \ .
\]
Note that if there is no regularization, then $\eta_j = -\nabla_j F(w)/\beta = g_j/\beta$ and this is the case we analyzed in the main body of the paper.
In the general case, by first order optimality conditions for the above one-dimensional convex optimization problem, we have
\begin{equation*}
	g_j + \beta \eta_j + \nu_j  = 0
\end{equation*}
where $\nu_j$ is a subgradient of $r$ at $w_j + \eta_j$. That is, $\nu_j \in \partial r(w_j + \eta_j)$. This implies that
\begin{equation*}
	r(w_j + \eta_j) - r(w') \le \nu_j(w_j + \eta_j - w')
\end{equation*}
for any $w'$.

We first calculate the expected change in objective function following the Shotgun analysis. We will use $w_b$ to denote $w_{j_b}$ (similarly for $\nu_b$, $g_b$)

\begin{align*}
\E{ f(\w') - f(\w) } &= P \Es{b}{ \eta_b g_b + \frac{\beta}{2} (\eta_b)^2 + r(w_b + \eta_b) - r(w_b)} \\
&\quad + \frac{\beta}{2} P(P-1) \Es{b \neq b'}{ \eta_b \cdot \eta_{b'} \cdot A_{j_b}^T A_{j_{b'}} } 
\end{align*}

Define the $B \times B$ matrix $M$ (depends on the current iteration) with entries $M_{b,b'} = A_{j_b}^T A_{j_b}$. Then, using
$r(w_b + \eta_b) - r(w_b) \le \nu_b \eta_b$, we continue
\begin{align*}
&\le \frac{P}{B} \left[ \eta^T g + \frac{\beta}{2} \eta^T \eta + \nu^T \eta \right] \\
&\quad +\frac{\beta P(P-1)}{2 B (B-1)} \left[ \eta^\top M \eta - \eta^T \eta \right] 
\end{align*}
Above (with some abuse of notation), $\eta$, $\nu$ and $g$ are $B$ length vectors with components $\eta_b$, $\nu_b$ and $g_b$ respectively.

Our generalization of Shotgun's $\rhoblock$ parameter is
\[
	\rhoblock = \max_{M\in \mathcal{M}} \rho(M)
\]
where $\mathcal{M}$ is the set of all $B \times B$ submatrices obtainable from $\X^T \X$ by selecting exactly one index from each of the $B$ blocks.

So, we continue
\begin{align*}
&\le \frac{P}{B} \left[ \eta^T g + \frac{\beta}{2} \eta^T \eta - g^T \eta - \beta \eta^T \eta \right] \\
&\quad +\frac{\beta P(P-1)}{2 B (B-1)} (\rhoblock - 1) \eta^T \eta 
\end{align*}
where we used $\nu = -g - \beta \eta$.

Simplifying we get
\[
\E{f(\w') - f(\w)} \le \frac{P\beta}{2B} \left[ -1 + \epsilon \right] \|\eta\|_2^2
\]
where
\[
	\epsilon = \frac{(P-1)(\rhoblock-1)}{(B-1)}
\]
should be less than $1$.

Now note that
\[
\|\eta\|_2^2 = \sum_{b} |\eta_{j_b}|^2 = \| \eta \|_{\infty,2}^2 \ .
\]
where the ``infinity-2'' norm $\|\cdot\|_{\infty,2}$ of a $p$-vector is, by definition, as follows: take the $\ell_\infty$ norm within a block
and take the $\ell_2$ of the resulting values. Note that in the second step above, we moved from a $B$-length $\eta$ to a $p$ length $\eta$.

This gives us
\begin{equation}\label{eq:keyaccbound}
	\E{f(\w') -f(\w)} \le -\frac{(1-\epsilon)P\beta}{2 B} \| \eta \|_{\infty,2}^2 \ .
\end{equation}

From the results in~\cite{Dhillon2011} we know that $f(\w) - f(\w^\star) \le C \|\eta\|_{\infty}$ where the constant $C$
depends on the function $F$ (e.g. its smoothness and Lipschitz constants) and the maximum value $\|\w - \w^\star\|_1$ can take over the course of the algorithm. Because
$\|\eta\|_\infty \le \|\eta\|_{\infty,2}$, plugging this into~\eqref{eq:keyaccbound}, we get
\[
	\E{f(\w') - f(\w)} \le -\frac{(1-\epsilon)P\beta}{2 B C } (f(\w)-f(\w^\star))^2 \ .
\]
Defining the accuracy $\alpha_k = F(w_k) - F(w^\star)$, we translate the above into the recurrence
\[
	\E{ \alpha_{k+1} - \alpha_{k} } \le -\frac{(1-\epsilon)P\beta}{2 B C } \E{\alpha_k^2}
\]
and by Jensen's we have $(\E{\alpha_k})^2 \le \E{\alpha_k^2}$ and therefore
\[
	\E{\alpha_{k+1}} - \E{\alpha_{k} } \le -\frac{(1-\epsilon)P\beta}{2 B C } (\E{\alpha_k})^2
\]
which solves to (upto a universal constant factor)
\[
	\E{\alpha_k} \le \frac{2 B C}{(1-\epsilon)P\beta} \cdot \frac{1}{k} \ .
\]

\end{document}